\documentclass[10pt,twocolumn,letterpaper]{article}

\usepackage{cvpr}
\usepackage{times}
\usepackage{epsfig}
\usepackage{graphicx}
\usepackage{amsmath}
\usepackage{amssymb}
\usepackage{multirow}

\usepackage{makecell}

\usepackage[pagebackref=true,breaklinks=true,letterpaper=true,colorlinks,bookmarks=false]{hyperref}

\cvprfinalcopy % *** Uncomment this line for the final submission

\ifcvprfinal\fi
\begin{document}

%%%%%%%%% TITLE
\title{Sem-LSD: A Learning-based Semantic Line Segment Detector}

%\author{Yi Sun \qquad Xushen Han \qquad Kai Sun \\
%Alibaba AI Labs\\
%Alibaba, Hang Zhou, China\\
%{\tt\small {tangyang.sy,xushen.hxs,sk157164}@alibaba-inc.com}

\author{Yi Sun, Xushen Han, Kai Sun, Boren Li, Yongjiang Chen, and Mingyang Li\\
	% {\tt\small \{sunyi324, xushen.hxs, sk157164, boren.lbr, yongjian.cyj\}@alibaba-inc.com}
	% For a paper whose authors are all at the same institution,
	% omit the following lines up until the closing ``}''.
	% Additional authors and addresses can be added with ``\and'',
	% just like the second author.
	% To save space, use either the email address or home page, not both
	% \and
	% Mingyang Li\\
	% {\tt\small mingyangli009@gmail.com}
}

\maketitle
%\thispagestyle{empty}
%%%%%%%%% ABSTRACT
\begin{abstract}
	In this paper, we introduces a new type of line-shaped image representation, named semantic line segment (Sem-LS) and focus on solving its detection problem.
	Sem-LS contains high-level semantics and is a compact scene representation where only visually salient line segments with stable semantics are preserved. 
	Combined with high-level semantics, Sem-LS is more robust under cluttered environment compared with existing line-shaped representations. 
	The compactness of Sem-LS facilitates its use in large-scale applications, such as city-scale SLAM (simultaneously localization and mapping) and LCD (loop closure detection).
	Sem-LS detection is a challenging task due to its significantly different appearance from existing learning-based image representations such as wireframes and objects.
	For further investigation, we first label Sem-LS on two well-known datasets, KITTI and KAIST URBAN, as new benchmarks. 
	Then, we propose a learning-based Sem-LS detector (Sem-LSD) and devise new module as well as metrics to address unique challenges in Sem-LS detection. 
	Experimental results have shown both the efficacy and efficiency of Sem-LSD. 
	Finally, the effectiveness of the proposed Sem-LS is supported by two experiments on detector repeatability and a city-scale LCD problem.
	Labeled datasets and code will be released shortly.
\end{abstract}

%%%%%%%%% BODY TEXT
\section{Introduction} \label{sec:intro}
Image representations play a fundamental role in many computer vision tasks, such as simultaneous localization and mapping (SLAM) \cite{slam_orbslam,slam_salient_feature_slam}, place recognition \cite{placerecognition_featurematching, placerecognition_context_based}, and motion tracking \cite{tracking_pointfeatures, tracking_objinimgsequence}. 
Line-shaped representations have been under active research in recent years \cite{slam_plslam,slam_linefeaturesANDrollingshutter,lineshaped_wireframe2018manmadeenv} to tackle the inherent challenges posed by conventional point representations, such as their difficulties in matching surfaces with repetitive patterns or those without texture. 
Meanwhile, endowing arbitrary lines with high-level semantic attributes is also of increased interests in research communities \cite{lineshaped_SLNet,lineshaped_horizonsky_lines}.
High-level semantics is shown to enhance the robustness of image representations \cite{LC_SemanticVisualLocalization,LC_segmentation} against frequently encountered challenging conditions in urban environments including occlusions, moving objects, seasonal and weather differences, illumination changes, and so on.
However, to the best of our knowledge, there is no existing works that combines high-level semantics and visually salient line segments in images.
Though objects with high-level semantics in urban streets such as road markings and street lambs can be stably detected across different conditions, e.g., seasons, illuminations, view point, and etc. via modern CNN based methods \cite{slam_overfeat}, they are too sparse to represent images.
Therefore, in this paper we focus on defining a new type of line-shaped image representation, Semantic Line Segment (Sem-LS), and solving its detection problem. 
%However, to the best of our knowledge, there is no existing works that combines high-level semantics and visually salient line segments in images.
%Given the fact that semantic features in urban streets such as road markings, street lambs, and vehicles can be stably detected across different conditions, e.g., seasons, illuminations, view point, and etc. via modern CNN based methods \cite{slam_overfeat}, 

%\textcolor{red}{However, objects are too sparse to be used as image representations.}

\begin{figure}[!t]
	\begin{center}
		\hspace{0cm}
		\includegraphics[width=0.8\linewidth]{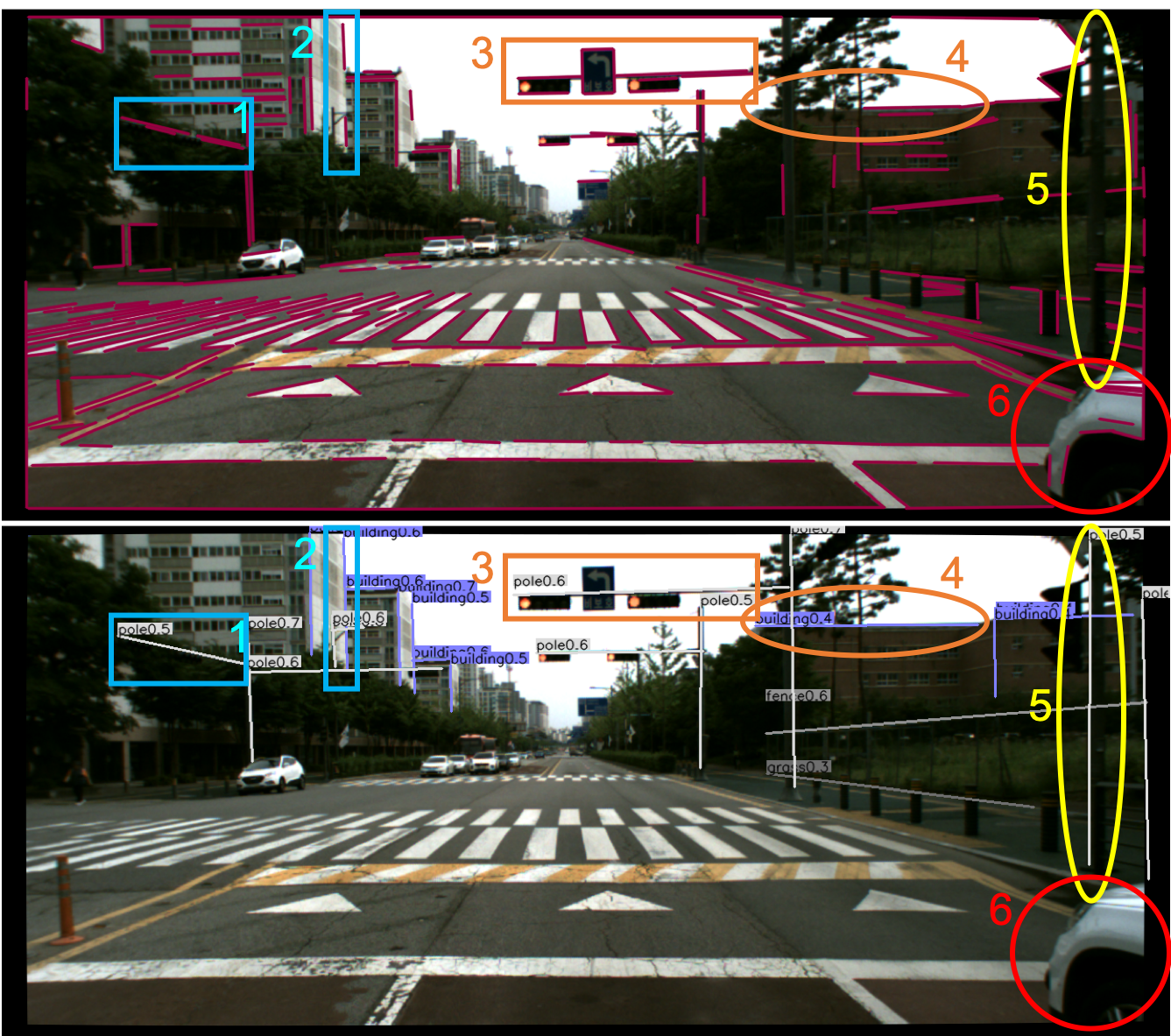}
	\end{center}
	\vspace{-10pt}
	\caption{Comparison between LSD \cite{lineshaped_LSD} (top) and proposed Sem-LSD (bottom). 1,2: Complete line segment detected as fragments by LSD; 3,4: Occluded line segments; 5: Line segment with low pixel-level gradient that is missed by LSD; 6: Unstable line segments detected by LSD. }
	\label{fig:motivation}
\end{figure}

A line segment is said to be a Sem-LS only when the two following conditions are both satisfied. 
Firstly, a Sem-LS must define \textit{straight boundaries} of a complete and visually salient object or structure. 
Secondly, a Sem-LS must have stable semantic meaning. 
Compared with existing line-shaped representations such as LSD results \cite{lineshaped_LSD} and wireframes \cite{lineshaped_wireframe2018manmadeenv}, 
Sem-LS captures only those complete and salient line segments and thus has more consistent appearance under image occlusions and aforementioned challenging conditions. 
Moreover, combined with high-level semantics, it is straight-forward to filter unstable regions such as moving objects. 
Additionally, Sem-LS provides a much lower density compared with existing line-shaped representations which largely facilitates its efficiency in computation and storage, especially for city-scale applications. 
The sparsity of Sem-LS is shown to be sufficiently dense by providing robust image matching in our experiment on loop closure detection (LCD).
%problem which acts as an example application of Sem-LS as well.
Figure \ref{fig:motivation} shows examples of Sem-LS and LSD results which illustrate the advantages of our proposed Sem-LS over existing line-shaped representations. 

Because of the large difference between definitions of Sem-LS and existing line segments, the detection of Sem-LS cannot be easily handled by existing detectors. 
LSD detects line segments based on pixel-level gradients which may not align with Sem-LS as shown in Figure \ref{fig:motivation} example 5. 
Wireframe parsers requires a joint encoding of line segments and their junctions, which are intentionally avoided by Sem-LS due to the weaker robustness against conditions such as changing illuminations, seasons, and occlusions. 
Furthermore, compared with objects, Sem-LS are line segments which vary significantly in terms of appearances and thus existing detectors cannot be directly applied. 

To this end, we develop an end-to-end learning-based Sem-LS detector (Sem-LSD) with novel Sem-LS encodings, customized modules, and modified metrics to tackle challenges posed by the unique definition of Sem-LS. 
Specifically, in adaptation to Sem-LS, we design two encodings, namely AngMidLen and LineAsObj (line as object). 
AngMidLen encodes Sem-LS by angle against positive x-axis, mid-point coordinates, and length while LineAsObj encodes a Sem-LS by its bounding box with an additional direction attribute. LineAsObj also enables a general encoding for both Sem-LS and objects. 
In terms of detector design, a customized Bi-Atrous Module (BAM) is proposed to handle Regions of Interest (RoI) with extreme width and height ratio of Sem-LS by covering large vertical and horizontal receptive fields simultaneously. 
As for evaluation, a modified metric Angle-Center-Length (ACL) is proposed in substitute of IoU to better quantize different overlapping situations between Sem-LS. 
Finally, to train and benchmark Sem-LSD, we labeled two new datasets based on KITTI \cite{dataset_kitti} and KAIST URBAN \cite{dataset_kaisturban} with totally over 540,000 annotations. Both datasets are outdoor city street views.

\begin{figure*}[!t]
	\begin{center}
		\hspace{0cm}
		\includegraphics[width=0.9 \linewidth]{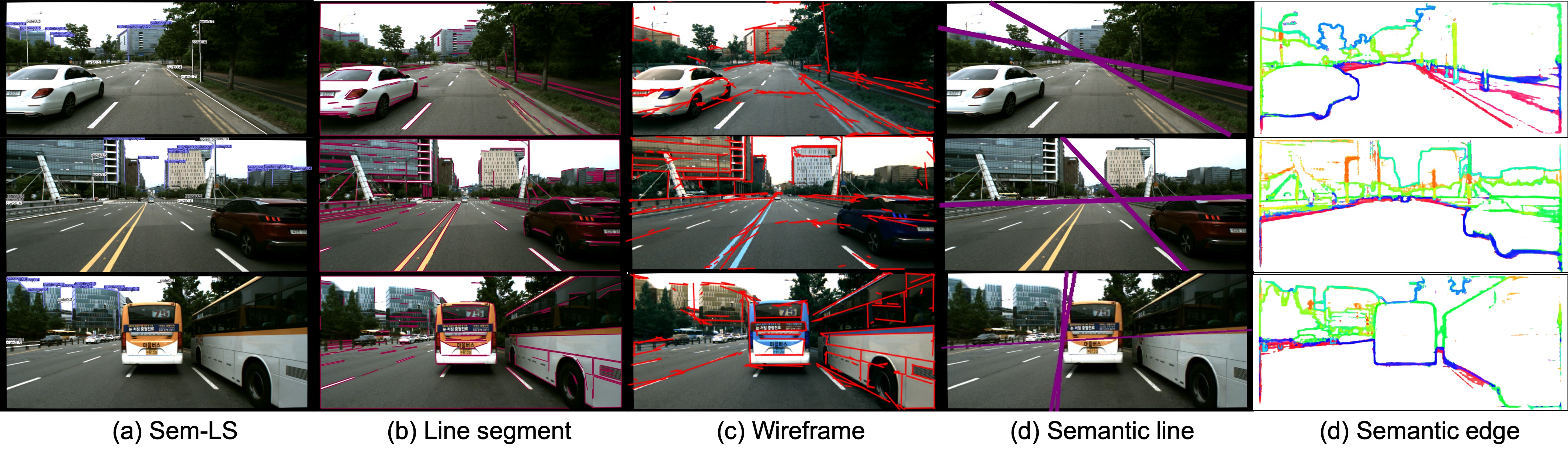}
	\end{center}
	\vspace{-10pt}
	\caption{Examples of different line-shaped local feature representations.
		(a) Sem-LS; (b) Line segments by LSD\cite{lineshaped_LSD}; (c) Wireframes by AFM \cite{lineshaped_afm}; (d) Semantic lines by SLNet \cite{lineshaped_SLNet}; (e) Semantic edges by CASENet \cite{segmentation_casenet}.
	}
	\label{fig:diff_lines}
\end{figure*}

Ablation study is conducted to demonstrate the efficacy of various Sem-LSD designs. 
In terms of efficiency, a light-weighted Sem-LSD is shown to run at over 160 fps on an commercial GPU and near real-time on an end device with little performance drop. 
We also discuss the possibility to leverage the general encoding based on LineAsObj to implement a unified Sem-LSD which detects Sem-LS and objects simultaneously. Such unified Sem-LSD is shown to perform comparatively with single-tasked detector. It hence doubles the detection efficiency and largely benefits real-world deployments with limited computational power. 

Last but not the least, We also conduct experiments on repeatability by applying random affine transformation on test images. Sem-LSD is shown to outperform by $15\%$ over the state-of-the-art line segment detector. 
A Sem-LS-based LCD algorithm is also implemented in order to provide an example of possible real-world application. 

To better benefit the community, we make the two datasets and code for our Sem-LSD publicly available. 

\noindent Our main contributions are three folds:

\noindent\textbf{1.}$\;$ We define a new learning-based line-shaped image representation, named Sem-LS, which capture complete and visually salient line segments with stable semantics.
%which is particularly suitable for outdoor city-scale applications. 

\noindent\textbf{2.}$\;$ We propose Sem-LSD with novel encodings, customized detector module, and evaluation metric to handle unique challenges in Sem-LS detection task. Two new labeled datasets are provided as benchmarks to support Sem-LS related researches. 

\noindent\textbf{3.}$\;$ We conduct extensive experiments to demonstrate the efficacy and efficiency of Sem-LSD. Sem-LSD is shown to provide better robustness by outperforming $15.08\%$ in repeatability and $16.01\%$ in LCD recall. 

%-------------------------------------------------------------------------
\section{Related work} \label{sec:related_work}

%Using local features as image representation has been well-studied for decades. 
%Keypoint descriptor extraction, object detection, and semantic segmentation have been well-studied for decades. 
%We review these tasks regarding their applications in robotics and autonomous driving so as to highlight the necessity of our proposed Sem-LS. 
%Existing work on line-shaped components are also reviewed to show the unique characteristics of Sem-LS. 

%\subsection{Keypoint descriptors}
\vspace{2pt}\noindent\textbf{Local point features.}$\;$
Using local point features as image representation has been well-established for more than a decade. 
Various features have been proposed including ORB \cite{descriptor_orb}, SIFT \cite{descriptor_sift}, SURF \cite{descriptor_surf}, FREAK \cite{descriptor_freak}, etc. 
Though enabling applications such as image retrieval and visual SLAM, the lack of high-level semantics makes them less robust in complex environment such as occlusions, seasonal differences, and illumination changes \cite{LC_CNN15_GaoXiang,LC_CNN17}.
Moreover, surfaces with repetitive patterns and those without textures remain challenging for point representations. 
%Moreover, the high density of point features require considerable storage space which can be preventive in city-scale applications. 

\vspace{2pt}\noindent\textbf{Existing line-shaped image representations.}$\;$
There are two types of existing line-shaped representations, namely, line segments \cite{lineshaped_LSD,lineshaped_Hough} and wireframes \cite{lineshaped_wireframe2018manmadeenv,lineshaped_afm,lineshaped_ppgnet}.
Line segments are extracted based on low-level features such as pixel-level gradients while
wireframes are consist of line segments and their junctions designed to encode scene geometry and object shapes. 
Similarly to point representations, these line-shaped representations do not capture high-level semantics and thus are less robust against challenging conditions mentioned in Section \ref{sec:intro}.
Furthermore, being scatter fragments, line segments and wireframes are more vulnerable to moving objects and occlusions. Their density also leads to high storage costs which can be prohibitive for city-scale applications.
Figure \ref{fig:diff_lines}(a)-(c) show examples of Sem-LS, LSD \cite{lineshaped_LSD} and wireframes \cite{lineshaped_afm} parsed by AFM.

\vspace{2pt}\noindent\textbf{Semantic lines and semantic edges.}$\;$
Semantic lines are proposed in \cite{lineshaped_SLNet} to define arbitrary image layouts whereas 
semantic edges \cite{segmentation_casenet, segmentation_steal} are extensions of semantic segmentation to detect edges and boundaries in between different semantic categories. 
Although with conceptually similarity, they target on different problems and applications.
Semantic lines capture highly sparse image layouts such as skylines that are not suitable as image representations. 
Semantic edges can not be parametrically represented in both 2D and 3D spaces, and thus not suitable in our use cases. 
Figure \ref{fig:diff_lines}(d)-(e) show examples of semantic lines and semantic edges.

\vspace{2pt}\noindent\textbf{Convolutions with irregular reception fields.}$\;$
Ordinary convolutions are designed to extract features from squared reception fields. In order to adapt to different shaped and scaled RoI, deformable convolutions \cite{network_deformable} and atrous convolutions \cite{network_dilatedconv} are proposed. 
By learning additional offsets for each kernel, deformable convolutions are able to cover arbitrary-shaped as well as adaptive-shaped reception fields which makes them suitable for non-squared RoI. 
On the other hand, atrous convolutions apply dilated rate in kernels so as to enlarge the scale of reception field without computational overheads. 
However, none of such convolutions is explicitly designed for extracting features from line-shaped RoI. 

%\subsection{Objects and segmentations}
%\vspace{2pt}\noindent\textbf{Objects and segmentations.}$\;$
%Objects and segmentations are ``things'' and ``stuff'' of an image \cite{panoptic_segmentation} which can also be taken as image features. Compared with keypoint descriptors, they convey semantics and are commonly deployed to help separate stable and unstable keypoints \cite{slam_salient_feature_slam,slam_dyna_slam,depth_estimation_obj_det}.
%Though successfully supporting applications such as SLAM in indoor scenes \cite{slam_object_slam2016,slam_indoor_obj}, static objects receive little attention in outdoor cases due to the sparsity. 
%Semantic segmentations are adopted similarly \cite{LC_segmentation,LC_SemanticVisualLocalization}.
%They are normally computationally heavy and their resultant masks cost much in parameterization and offline storage.
%but it is unrealistic to store the entire mask for large-scale applications. % such as robotic pre-built map.

%\subsection{Semantic edge segmentation}

%------------------------------------------------------------------------
\section{Sem-LS datasets} \label{sec:dataset}

In this section, we present two new labeled Sem-LS datasets, namely, KITTI-Sem-LS and KAIST-Sem-LS.
We labeled all images from KITTI object detection task \cite{dataset_kitti} with 7,481 for training and 7,518 for testing.
As for KAIST-Sem-LS, we selected the 39th sequence recorded in Korean city Pankyo which has most of the challenging conditions mentioned in Section \ref{sec:intro}. We evenly sampled one fifth of the frames to collect 3,729 images and split into training and testing datasets with a $85\%/15\%$ ratio.  
%We split KAIST-Sem-LS into training and testing with a $85\%/15\%$ ratio. 
%It is recorded at 10 fps and we extract 2 frames every second to collect 3,729 images. 
%Figure \ref{fig:kaist_kitti_datasets} (a) is examples from KAIST-Sem-LS and Figure \ref{fig:kaist_kitti_datasets} (b) is from KITTI-Sem-LS.
We define 14 categories and the statistics of the 4 most representative categories are listed in Table \ref{tab:dataset_stats}.
For each Sem-LS, we label its two endpoints and the category.
There are totally 465,950 and 77,779 labels in KITTI-Sem-LS and KAIST-Sem-LS, respectively. KITTI-Sem-LS has 31.07 labels per image on average while KAIST-Sem-LS has 22.71. 
%Figure \ref{fig:kaist_kitti_datasets} shows examples from KAIST-Sem-LS (left column) and KITTI-Sem-LS (right column).

\begin{table}
	\begin{center}
		\begin{tabular}{c|c|c}
			\hline
			 & KITTI-Sem-LS & KAIST-Sem-LS  \\
			\hline\hline
			building & 65,576 & 22,898  \\
			\hline
			pole & 121,691 & 24,622 \\
			\hline
			curb & 54,454 & 6,894 \\
			\hline
			grass & 32,121 & 1,174 \\
			\hline
			Total & 465,950 & 77,779 \\
			\hline
			labels/image & 31.07 & 22.71 \\
			\hline
		\end{tabular}
	\end{center}
	\vspace{-5pt}
	\caption{Statistics of KAIST-Sem-LS and KITTI-Sem-LS annotations}
	\label{tab:dataset_stats}
\end{table}

%\textbf{Version 1: }
%To improve the precision of manual labeling, we provide a gradient-based semi-auto toolkit for our labelers. For each line segment labeled by the labeler, the toolkit will automatically search for near-by LSD results. If any of the LSD result is similar enough to the labeled line segment in terms of angle, center position, and length, the toolkit will also display the one with the highest similarity as candidate for this labeled line segment. The labeler will then be suggested to select the LSD one instead of the original one. In this way, we can alleviate human labeling errors. 

%\textbf{Version 2:}
To improve the precision of manual labeling, we adopt a gradient-based correction. For each labeled Sem-LS $S$, LSD \cite{lineshaped_LSD} is deployed to find if there exists a detection $S_{lsd}$ that satisfies $Overlap(S, S_{lsd}) > 0.95$, 
where $Overlap()$ is some function to measure the level of overlap between line segments. We adopt our proposed ACL as the $Overlap()$ in our implementation.
%where $ACL()$ is our newly proposed metric measuring the overlap between line segments. 
If such $S_{lsd}$ exists, the original $S$ will be replaced by $S_{lsd}$. 
The rationale behind is that in some cases, visually salient line segments do have the largest pixel-level gradient.
%The rationale behind is that in some cases, semantic gradient aligns with pixel-level gradient. 
%A strict thresholding is applied to avoid wrongly corrections when semantic gradients actually differ from pixel-level gradients. 
Therefore, gradient-based line segment can be used to correct small human labeling drifts. 
A strict threshold at $0.95$ is applied to avoid wrong corrections when $S$ does not align with pixel-level gradient. 

We evaluate the labeling error before and after the correction by reporting a line-segment-based projection error from sampled image triplets. 
%Specifically, we build image triplets with images from left and right camera of sequence 39 and their matched left camera image in terms of GPS distance of sequence 28. 
Specifically, we build each triplet with three selected images from KAIST-Sem-LS. 
One of the image is the left-camera image $I_{left}$. 
The other two are the correspondent right-camera image $I_{right}$ and the preceding left-camera image $I_{pre}$ with distance $D$ to the $I_{left}$. 
The distance is calculated by ground truth pose provided by the KAIST URBAN dataset. In our experiment, we use $D=8$ meters.
To quantize labeling error, we project all correspondent Sem-LS in $I_{left}$ and $I_{pre}$ into the 3D space and then re-project the lines in 3D space back to $I_{right}$. 
Endpoint-to-line distances are calculated between the labeled Sem-LS in $I_{right}$ and the correspondent re-projected line. 
We report the averaged error over all image triplets.
The projection error on original labels is 1.91 pixels and the refined labels achieves a 1.73-pixel project error. 
%Correspondent Sem-LSs from sequence 39 two cameras are projected to 3D space then re-projected onto the third image space. Labeling error is calculated based on endpoint-to-line distances between the re-projected line segment and its ground truth label.
%We average such errors over all valid image triplets.
%Projection error on original labels is 5 pixels while it achieves 2.5 pixels on refined labels. 
%Such error defines the upper bound of geometric precision for the predicted Sem-LSs. 
An error less than 2 pixels is reasonable for applications such as visual perception and scene understanding. 

%We make no assumptions such as Manhattan world assumption on our labels. 
%For pillar-shaped poles and trees with two paralleled or nearly paralleled segments, we did our best to label both of them. For those poles or trees that are remote to the camera, we labeled the one close to the road side. 

%------------------------------------------------------------------------
\section{Sem-LS Detection} \label{sec:Sem-LSd}

\begin{figure*}[!th] 
	\begin{center}
		\hspace{0cm}
		\includegraphics[width=0.8\linewidth]{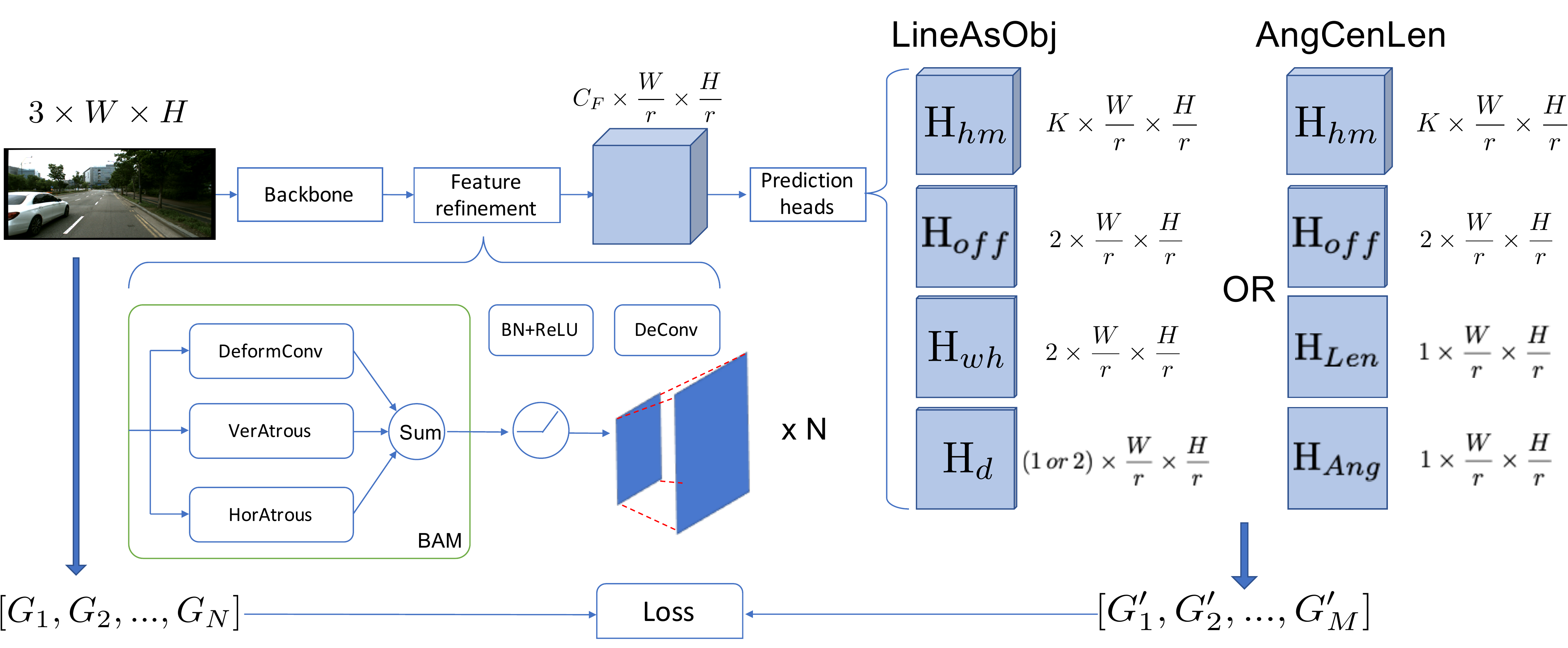}
	\end{center}
	\vspace*{-10pt}
	\caption{General workflow of our proposed Sem-LS detector.}
	\label{fig:model_arch}
\end{figure*}

\begin{figure}[!h]
	\begin{center}
		\hspace{0cm}
		\includegraphics[width=0.9\linewidth]{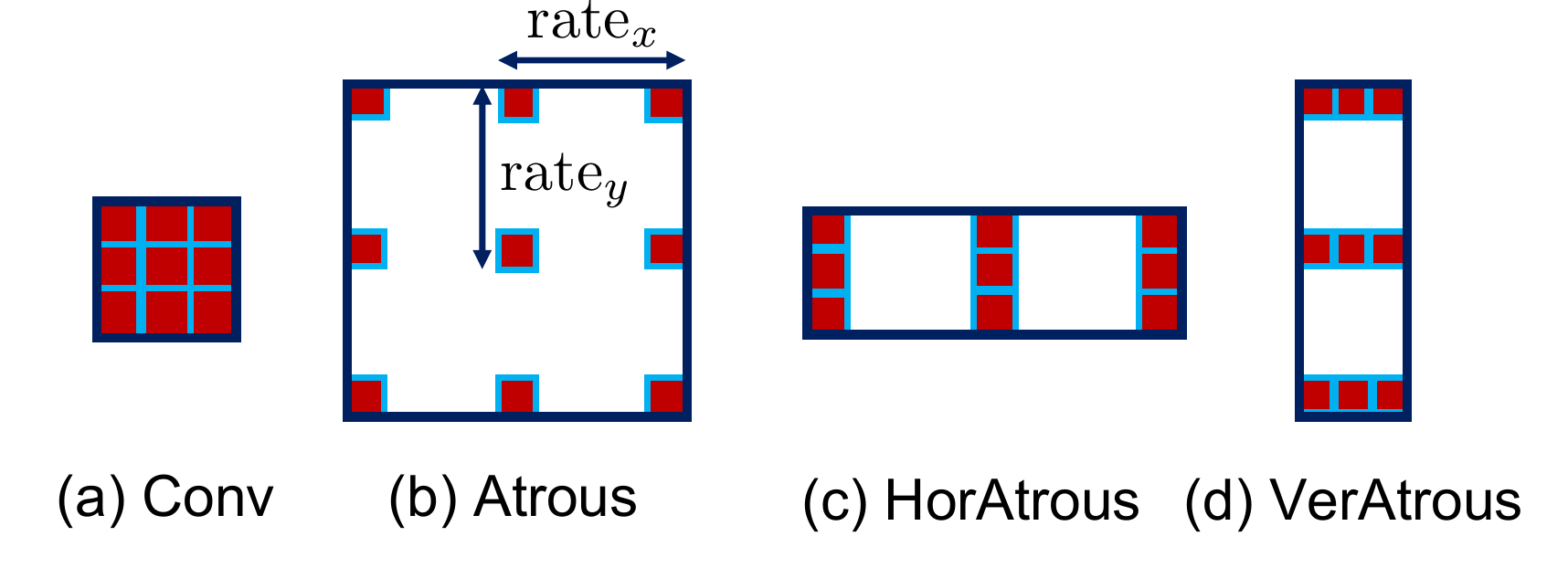}
	\end{center}
	\vspace{-10pt}
	\caption{An illustration of different convolutions with kernel size $3\times3$. (a): an ordinary convolution; (b): a regular atrous convolution with $rate_x=rate_y$; (c) Proposed VerAtrous convolution with $rate_y > rate_x$; (d) Proposed HorAtrous convolution with $rate_x > rate_y$.} %(e) A BAM consist of deformable convolution, VerAtrous, and HorAtrous in parallel.}
	\label{fig:AtrousModule}
\end{figure}

In this section, we describe our proposed solution to Sem-LS detection task including Sem-LS encodings, Sem-LSD architectures, and loss function designs. 
%In this section, we first introduce the proposed Sem-LS representation ``line as object''. Then we describe the detector Sem-LSD with customized layer BAM. 

\subsection{Sem-LS encodings} \label{sec:representation}
%\subsection{Sem-LS representation - ``Line as Object''} \label{sec:representation}
%\vspace{2pt}\noindent\textbf{Sem-LS representation}$\;$
Existing wireframe parsing models predict junction positions and line heat maps then combine them to get the final detections \cite{lineshaped_wireframe2018manmadeenv,lineshaped_endtoend}. 
Such junction-line segment encoding is not suitable for Sem-LS since Sem-LS are not necessarily intersect with each other. 
%Their end points may even be ambiguous under some circumstances. 
To this end, we propose two different Sem-LS encodings, AngMidLen and LineAsObj, with Sem-LS geometry encoded differently.
AngMidLen encodes the geometry of Sem-LS by its Angle against positive x-axis, Mid-point coordinates, and Length. 
LineAsObj, on the other hand, follows the observation that each Sem-LS can be seen as one of the diagonals of its minimum bounding box.
It encodes each Sem-LS similarly as bounding boxes using an extra attribute, direction. The direction is either left-top to right-bottom or left-bottom to right-top. 

%To find correct centers and angles are more important than accurate end point locations. 
%We follow the observation that each Sem-LS can be seen as one of the diagonals of its minimum bounding box to present ``line as object'' which represents each Sem-LS similarly with bounding boxes using an extra attribute, direction. ``Line as object'' also enables a general representation for both Sem-LS and object.
LineAsObj also enables a general encoding for both Sem-LS and object.
Specifically, let $S$ denotes a Sem-LS and $B$ denotes its minimum bounding box, we define the general encoding as $G=B\cup S=(\mathrm{p}, \mathrm{d_g}, \mathrm{k})$, 
where $\mathrm{p}$ denotes the encoded geometry of $G$, $\mathrm{d_g}$ is the direction and $\mathrm{k}$ is the index of semantic category.
Hence, we have object encoded as $B = (G|\mathrm{d_g}=2)$ and Sem-LS as $S = (G|\mathrm{d_g}\in \left\{0, 1\right\})$. 
$\mathrm{d_g}=0$ represents Sem-LS from left-top to right-bottom and $\mathrm{d_g}=1$ for the other direction.
Geometry is encoded by $\mathrm{p}=(x_c,y_c,w,h)$, where $(x_c,y_c)$ is the center point coordinates and $w$, $h$ are width and height of the bounding box.

\subsection{Sem-LS Detector (Sem-LSD)}
%\subsubsection{Network design}
\noindent\textbf{Network design.}$\;\;$
%Upsampling before prediction heads is widely adopted in semantic segmentation since U-Net \cite{segmentation_unet}. 
%The works on stacked hourglass networks \cite{network_stackedhourglass} further promotes such hourglass-like structure to be adopted in object detectors \cite{obj_cornernet, obj_m2det, obj_tinydsod}.  With each feature map position representing one candidate of object keypoint (selected corner or center), anchor-free detectors such as \cite{obj_cornernet, obj_objectaspoint} are especially benefited.
The overall structure of Sem-LSD is shown in Figure \ref{fig:model_arch}. 
As anchor-based detector has limited width-height ratio on anchor boxes which does not fit vertical and horizontal line segments, we build Sem-LSD based on a recent anchor-free detector CenterNet \cite{obj_objectaspoint}. CenterNet is selected for its structural simplicity, top-level performance, and verified flexibility for multiple tasks. 
The hourglass-like structure of Sem-LSD includes a backbone for spatial feature extraction and a feature refinement network to a) aggregate multi-scale features; b) upsample intermediate feature maps to allow more prediction candidates \cite{obj_objectaspoint}.

Given an input image $I \in \mathbb{R}^{3\times W \times H}$ with $N$ ground truth Sem-LS $[G_1, G_2, ..., G_N]$, the final feature map $F\in\mathbb{R}^{C_F\times\frac{W}{r}\times\frac{H}{r}}$, where $r$ is output stride and is normally set to $2^k$, $k\in\mathbb{N}$. We set $r=4$ to be same with the literature \cite{pose_r4_00, pose_r4_01, network_stackedhourglass}. $C_F$ is number of channels for $F$. Predicted Sem-LS are denoted as $[G^\prime_1,G^\prime_2,...,G^\prime_M]$.
We omit batch size from dimension for simplicity. 

%\subsubsection{BAM}
\vspace{2pt}
\noindent\textbf{BAM (BAM).}$\;\;$
Atrous convolution is originally proposed for semantic segmentation task \cite{network_dilatedconv}. 
By adjusting dilation rate, it expands reception field without introducing extra computations. 

%In terms of Sem-LS detection, vertical and horizontal line segments are typically challenging due to its extreme scale. The significant variation in lengths also requires larger reception filed.
As Sem-LSD is designed based on existing object detectors, it is capable to handle rectangle RoI.
However, nearly vertical and horizontal line segments have RoI with extreme width and height ratios which cannot be effectively handled by existing convolutions. 

To address this challenge, we propose Vertical Atrous (VerAtrous) and Horizontal Atrous (HorAtrous) convolution. As shown in Figure \ref{fig:AtrousModule}, we set different dilation rates on vertical and horizontal directions. VerAtrous has $\mathrm{rate}_y > \mathrm{rate}_x$ whilst HorAtrous has $\mathrm{rate}_y < \mathrm{rate}_x$. VerAtrous is designed to provide a larger reception field along y-axis and a smaller one along x-axis to better fit those vertical or near-vertical line segments. HorAtrous is designed similarly for horizontal line segments. 

%We build BAM using a pair of VerAtrous and HorAtrous together with a deformable convolution \cite{network_deformable}.
We build BAM with three paralleled modules, namely, VerAtrous, HorAtrous, and a deformable convolution \cite{network_deformable}.
BAM is designed to extract and aggregate multi-scale and multi-aspect ratio features without changing feature map dimension. 
It is implemented right before the deconvolution layer (DeConv) in feature refinement network. 
All three modules in BAM are followed by batch normalization and ReLU activation.
Pixel-wise summation is applied to aggregate three feature maps as the output of BAM.
%We apply pixel-wise summation to aggregate feature maps from three paths. 
%Let $F_{ver}$, $F_{hor}$, and $F_{deform}$ denote the output feature maps from VerAtrous, HorAtrous, and deformable convolution, the module output feature map $F_{BAM}$ is calculated as follows:
%\vspace{-1pt}
%\begin{equation}
%F_{BAM} = \alpha F_{ver} + \beta F_{hor} + \eta F_{deform},
%F_{BAM} = F_{ver} + F_{hor} + F_{deform},
%\end{equation}
%where $\alpha=\beta=\eta=1$ for regular summation mode. For trainable weighted mode, $\alpha$, $\beta$, and $\eta$ are trainable variables with initial values at $\alpha = \beta = 0.3$ and $\eta = 0.4$. During training, they are restricted by $\alpha + \beta + \eta = 1$ and $\alpha,\beta,\eta\in(0,1)$.

%\subsubsection{Prediction heads}
\vspace{2pt}
\noindent\textbf{Prediction heads.}$\;\;$
%As shown in Figure \ref{fig:model_arch}, we introduce an additional direction head ($\mathrm{H}_d$) besides the original heat map head ($\mathrm{H}_{hm}$), width and height head ($\mathrm{H}_{wh}$), and offset head ($\mathrm{H}_{offset}$) in \cite{obj_objectaspoint}. 
%$\mathrm{H}_{d}$ is designed specifically for Sem-LS detection to predict the direction of all Sem-LSs. 
%It predicts the direction $\mathrm{D}=[0,1] ^ {2 \times \frac{W}{r} \times \frac{H}{r}}$ of all Sem-LSs.
As shown in Figure \ref{fig:model_arch}, we implement different prediction heads for AngMidLen and LineAsObj representations, respectively. 
$\mathrm{H}_{hm}$ and $\mathrm{H}_{off}$ are common heads that are deployed similarly as in \cite{obj_objectaspoint} to predict center point coordinates for positive detections. 
Specifically, $\mathrm{H}_{hm}$ predicts a heat map of center point $\hat{\mathrm{M}}\in[0, 1] ^ {K \times \frac{W}{r} \times \frac{H}{r}}$, where the number of channels $K$ equals to the number of semantic categories. 
For each ground truth center point $c\in\mathbb{R}^{2}$, its equivalent position on $F$ is calculated by $\widetilde{c}=\left\lfloor\frac{c}{r}\right\rfloor$. The ground truth heat map $M\in[0,1]^{K \times \frac{W}{r} \times \frac{H}{r}}$ is generated by assigning values according to a Gaussian kernel, $M_{kxy} = \mathrm{exp}(-\frac{(x-\widetilde{c}_x)^2 + (y-\widetilde{c}_y)^2 }{2\sigma_p^2})$, where $\sigma_p^2$ is a self-adaptive standard deviation with respective to $\mathrm{min}(w, h)$
%and $w$, $h$ are the width and height of ground truth Sem-LS centered at $(x, y)$. 
Only at center coordinate $(x,y)$ of ground truth line segments with category $k$, we have $M_{kxy}=1$.
$\mathrm{H}_{off}$ tries to recover the error $\mathrm{E}\in\mathbb{R}^{2\times \frac{W}{r} \times \frac{H}{r}}$ introduced by $r$ and the floor operation between $\widetilde{c}$ and $c$. 

With AngMidLen encoding, $\mathrm{H}_{Ang}$ and $\mathrm{H}_{Len}$ predict the angle and length of each Sem-LS. Both heads have output dimension at $1\times\frac{W}{r}\times\frac{H}{r}$.

In terms of LineAsObj, 
$\mathrm{H}_{wh}$ is implemented to predicts width and height $\mathrm{WH}\in{\mathbb{R}_{\geq0}}^{2\times \frac{W}{r} \times \frac{H}{r}}$ for bounding boxes centered at each feature map location. 
We introduce an additional direction head $\mathrm{H}_{d}$ to predict the directions of all Sem-LS. Depending on the loss function used, $\mathrm{H}_{d}$ outputs tensor at dimension $1\times\frac{W}{r}\times\frac{H}{r}$ or $2\times\frac{W}{r}\times\frac{H}{r}$. 
%Detailed explanations are provided below.

%\subsubsection{Loss function}
\vspace{2pt}
\noindent\textbf{Loss functions.}$\;\;$
%Loss function hence consists of four parts:
Loss function for AngMidLen ($L_{AML}$) and LineAsObj ($L_{LaO}$) are given as:
\begin{equation}
L_{AML} = L_{hm} + \lambda_{off}L_{off} + \lambda_{Ang}L_{Ang} + \lambda_{Len}L_{Len},
\end{equation}
\begin{equation}
L_{LaO} = L_{hm} + \lambda_{off}L_{off} + \lambda_{wh}L_{wh} + \lambda_{d}L_{d},
\end{equation}
where $L$ are losses from prediction heads and $\lambda$ are corresponding weights.
%where $L_{hm}$, $L_{wh}$, $L_{off}$, $L_{d}$ are loss from four prediction heads, respectively. $\lambda_{wh}$, $\lambda_{off}$, $\lambda_{d}$ are corresponding weights. 

For both representations, $L_{hm}$ calculates a per-pixel logarithmic loss:
\begin{equation}
L_{hm} = \frac{1}{N} \sum_{kxy} {M_t}^\delta (1-\hat{M}_t)^\gamma \log(\hat{M}_t),
\end{equation}
where $\delta$ and $\gamma$ are \textit{focusing} parameters \cite{obj_focalloss}. We set $\delta=4$ and $\gamma=2$ \cite{obj_cornernet} throughout our experiments.
$M_t$ and $\hat{M}_t$ are defined as:
\begin{equation}
M_t = \left\{\begin{matrix}
1, &\mathrm{if}\; M_{kxy} = 1 \\
1-M_{kxy}, &\mathrm{otherwise}
\end{matrix}
\right.,
\end{equation}
\begin{equation}
\hat{M}_t = \left\{\begin{matrix}
\hat{M}_{kxy}, &\mathrm{if}\; M_{kxy}=1\\ 
1-\hat{M}_{kxy}, &\mathrm{otherwise}
\end{matrix}\right..
\end{equation}
%As ground truth can be only $0$ or $1$, $L_{hm}$ calculates a per-pixel logarithmic loss. 
$L_{off}$ is calculated by masked L1 loss where only those ground truth positions are considered:
\begin{equation}
L_{off} = \frac{1}{N}M_{gt}\sum_{e=1}^2\sum_{xy} |\hat{E}_{exy} - E_{exy}|,
\end{equation}
where $M_{gt}$ is the ground truth mask defined as 
\begin{equation}
M_{gt} = \left\{\begin{matrix}
1, &\mathrm{if}\; \underset{K}{\mathrm{max}}M_{kxy}=1 \\
0, &\mathrm{otherwise}
\end{matrix}
\right..
\end{equation}

We also adopt masked L1 loss for $L_{Ang}$, $L_{Len}$, and $L_{wh}$.

Two different loss functions, namely, regression loss and classification loss, are implemented to learn directions with LineAsObj.
Regression loss is given by a masked L1 loss:
\begin{equation}
L_{d_{reg}} = \frac{1}{N}M_{gt}\sum_{xy}|\hat{D}_{xy} - {D}_{xy}|,
\end{equation}
and classification loss is designed as a masked cross-entropy loss as below:
\begin{equation}
L_{d_{cls}} = - M_{gt} log(\frac{\mathrm{exp}(\hat{D_{dxy}})}{\sum_{d=1}^2 \mathrm{exp}(D_{dxy})}).
\end{equation}

\section{Evaluation metric} \label{sec:acl}
\begin{figure}[!t]
	\begin{center}
		\hspace{0cm}
		\includegraphics[width=0.9\linewidth]{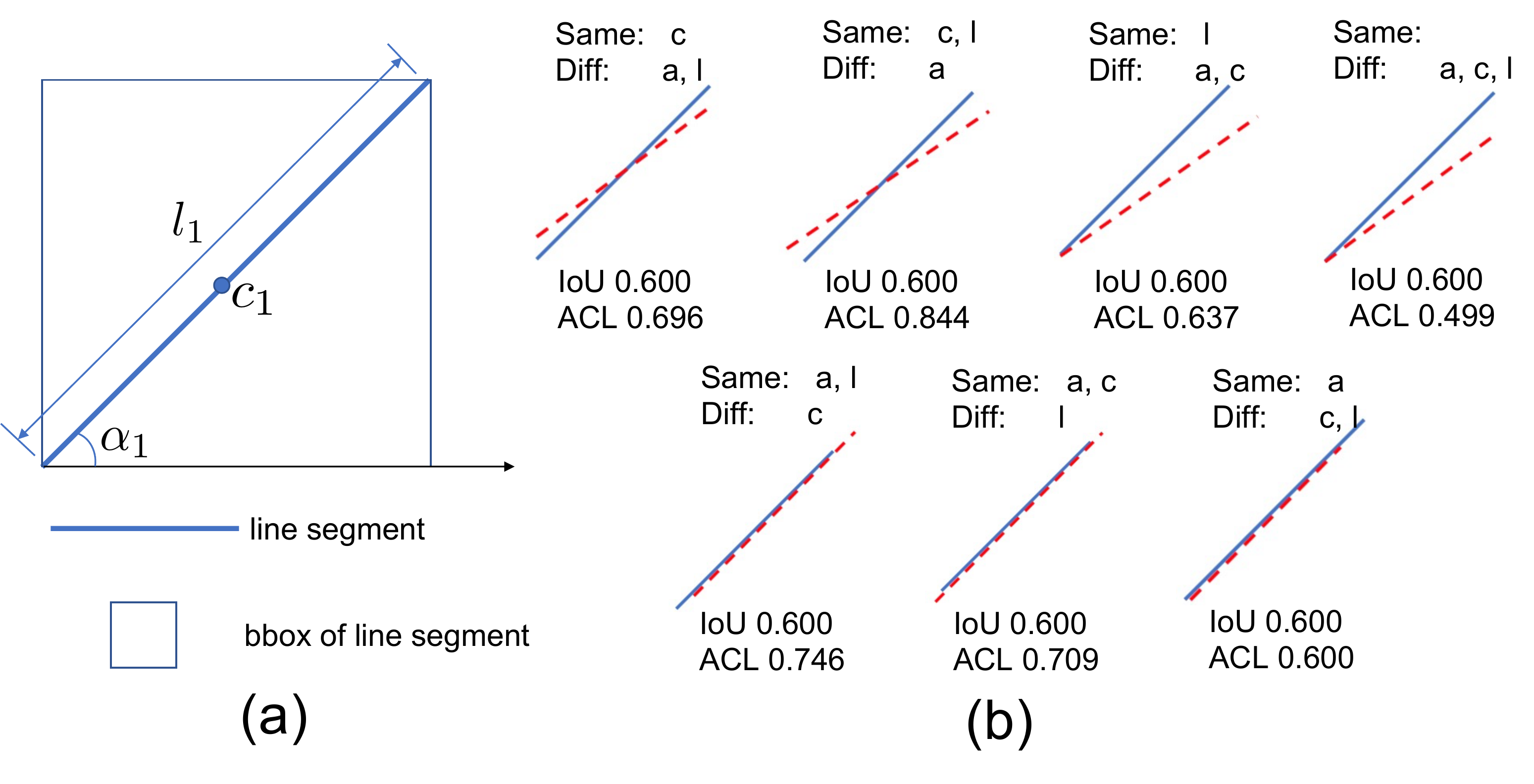}
	\end{center}
	\vspace{-10pt}
	\caption{(a) An example of a Sem-LS and its minimum bounding box. Its angle $\alpha$, center point $c$, and length $l$ are displayed. (b) Different overlapping cases of two line segments. All cases have the same IoU value (0.6) and different ACL values.}
	\label{fig:aclcases}
\end{figure}

%\subsection{Evaluation metric} \label{sec:acl}
%\vspace{2pt}\noindent\textbf{Evaluation metric.}$\;$
IoU is widely adopted as a metric to measure the overlapping of two object bounding boxes. 
As Sem-LS can be considered as the diagonal of its minimum bounding box, we may directly apply IoU to quantize the overlapping between two Sem-LS.
%As we have noted in Section \ref{sec:Sem-LSd}, Sem-LS can be considered as the diagonal of its bounding box. Therefore, we may directly apply IoU to calculate the overlapping level of two Sem-LS. 
However, since IoU considers only the position and size of boxes but not the width and height ratio, it is not optimized for Sem-LS. 
Width and height ratio decides the angle of Sem-LS which is a critical attributes to measure the detection accuracy.
Figure \ref{fig:aclcases}(a) shows an example of Sem-LS and its minimum bounding box.
We design ACL as a replacement for IoU to achieve desired overlapping measurement. ACL is calculated based on the difference between \textbf{A}ngles, \textbf{C}enter coordinates, and \textbf{L}engths of two line segments.

Specifically, given two Sem-LS $S_i$ represented by $[{x_c}_i, {y_c}_i, {w}_i, {h}_i, \mathrm{d_g}_i, \mathrm{k}_i]$, where $i \in \left\{1, 2\right\}$,
%Specifically, given two Sem-LSs $S_i$ represented by $[{x_s}_i, {y_s}_i, {x_e}_i, {y_e}_i, \mathrm{k}_i]$, where $({x_s}_i, {y_s}_i),\;({x_e}_i, {y_e}_i)$, $i\in[1,2]$ denote the coordinate of two end points. $\mathrm{k}_i$ denotes the semantic category index. 
%$[{x_L}_i, {y_L}_i, {x_R}_i, {y_R}_i, C]$, where $i \in [1, 2]$.
%Given two Sem-LSs $S_1$ and $S_2$, let $\alpha_1$, $c_1=(x_1, y_1)$, $l_1$ and $\alpha_2$, $c_2=(x_2, y_2)$, $l_2$ represent the angle against positive x-axis, the coordinates of line segment center point, and the length of 
ACL is then calculated as follows:
\begin{equation} \label{eq:acl}
ACL(S_1, S_2) = \left\{\begin{matrix}
Sim_\alpha \times Sim_c \times Sim_l ,\; if \; \mathrm{k}_1 = \mathrm{k}_2 \\ 
\\
0, \; \mathrm{otherwise}
\end{matrix}\right.,
\end{equation}
where ${Sim}_\alpha$, ${Sim}_c$, and ${Sim}_l$ are center coordinate similarity, length similarity, and angle similarity, respectively.

${Sim}_c$ is given by:
\begin{equation} \label{eq:simc}
{Sim}_c = 1 - \frac{\left\|c_1, c_2\right\|_2}{0.5 \times l_1},
\end{equation}
where $c_i = ({x_c}_i, {y_c}_i)$, $i\in\left\{1,2\right\}$ is center point coordinates of two Sem-LS.

${Sim}_l$ is given by:
\begin{equation} \label{eq:siml}
{Sim}_l = 1 - \frac{|l_1 - l_2|}{l_1},
\end{equation}
where $l_1$ and $l_2$ are length of two Sem-LS. 

${Sim}_\alpha$ is given by:
\begin{equation} \label{eq:simangle}
{Sim}_\alpha = 1 - \frac{|\alpha_1 - \alpha_2|}{90^\circ}, 
\end{equation}
where $\alpha$ is the angle against positive x-axis. 
%\begin{equation}
%\alpha_i = \left\{\begin{matrix}
%\frac{180^\circ}{\pi}\arctan\frac{{y_e}_i - {y_s}_i}{{x_e}_i - {x_s}_i}, \; if \; {x_s}_i \ne {x_e}_i \\ 
%\\
%90^\circ,\; otherwise
%\end{matrix}\right..
%\end{equation}

To further demonstrate the effectiveness of ACL, we show seven different overlapping cases in Figure \ref{fig:aclcases}(b) (best viewed in color). 
Each case has a unique combination of attributes ($\alpha$, $c$, and $l$) that are identical or different.
While all cases share the same IoU value (0.6), ACL returns different values. ACL value tends to be higher when more attributes are same and lower vise versa. This empirically verifies that ACL is more suitable to measure overlapping of line segments. 

Based on the proposed ACL, we modify mAP from MS COCO \cite{dataset_mscoco} to develop two new metrics, namely, $\mathrm{mAP}@0.5$ and $\mathrm{mAP}_3@0.5$, where $0.5$ is the minimum threshold for detection confidence.
Similar to mAP, $\mathrm{mAP}@0.5$ averages precisions over all categories. $\mathrm{mAP}_3@0.5$ considers three categories with the most occurrence, namely, pole, building, and curb. 
ACL is used instead of IoU.
We calculates precision over ten ACL values at $[0.5, 0.55, 0.6, ..., 0.95]$.

%------------------------------------------------------------------------
\begin{table*}[!t]
	\begin{center}
		\begin{tabular}{c|c|c|c|c|c|c|c|c}
			\hline
			\multirow{2}{*}{Models} & \multirow{2}{*}{LineAsObj} & \multirow{2}{*}{D-Cls} & \multirow{2}{*}{BAM51} & \multirow{2}{*}{BAM33} & \multicolumn{2}{c}{KAIST-Sem-LS} & \multicolumn{2}{c}{KITTI-Sem-LS} \\
			\cline{6-9}
			& & & & & $\mathrm{mAP}@0.5$ & $\mathrm{mAP}_3@0.5$ & $\mathrm{mAP}@0.5$ & $\mathrm{mAP}_3@0.5$ \\
			\hline\hline
			%LSD    & 0.0086 & --- \\
			%\hline
			baseline & & & & & 0.4742 & 0.6605 & 0.2607 & 0.6434 \\
			\hline
			Sem-LSD\_v0 & \checkmark & & & & 0.4559 & 0.6684 & 0.4103 & 0.6529 \\
			\hline
			Sem-LSD\_v1 & \checkmark & \checkmark & & & 0.4837 & 0.6591 & 0.4533 & \textbf{0.6680} \\
			\hline
			Sem-LSD\_BAM51 & \checkmark & \checkmark & \checkmark & & 0.4841 & 0.6562 & \textbf{0.4670} & 0.6532 \\
			\hline
			Sem-LSD\_BAM33 & \checkmark & \checkmark & & \checkmark & \textbf{0.5002} & \textbf{0.6805} & 0.4304 & 0.6494 \\
			\hline
		\end{tabular}
	\end{center}
	\caption{Results from Sem-LSD using different BAM with dla34 backbone on KITTI-Sem-LS dataset.}
	\label{tab:Sem-LSD-ablation_v1}
\end{table*}

\section{Experimentations} \label{sec:exp}

We present two parts of experimentations. 
The first one thoroughly evaluates the efficacy and efficiency of our proposed Sem-LSD on Sem-LS detection tasks. 
The other part demonstrates that Sem-LSD achieves better robustness under changing image appearances.

%\subsection{Implementation and results}
\vspace{2pt}\noindent\textbf{Sem-LSD evaluation.}$\;$
We first evaluate gains from different modifications on Sem-LSD in Table \ref{tab:Sem-LSD-ablation_v1}. DLA-34 \cite{network_dla} is adopted as backbone. 
We implement two types of BAMs, namely, BAM51 and BAM33. BAM51 has kernel size $1\times 5$ and $5\times 1$ for VerAtrous and HorAtrous, respectively. VerAtrous has $\mathrm{rate}_x=1$ and $\mathrm{rate}_y=2$ while HorAtrous has $\mathrm{rate}_x=2$ and $\mathrm{rate}_y=1$. To keep input dimension, we use a $(4,0)$ padding for VerAtrous and $(0,4)$ for HorAtrous. 
BAM33 has kernel size $3\times 3$ for both VerAtrous and HorAtrous. VerAtrous uses $\mathrm{rate}_x=1$ and $\mathrm{rate}_y=3$ with $(1,3)$ padding. HorAtrous has $\mathrm{rate}_x=3$ and $\mathrm{rate}_y=1$ with $(3,1)$ padding.
Deformable convolution has kernel size $3\times3$, stride 1 and padding 1.
We test all models on both KITTI-Sem-LS and KAIST-Sem-LS.
AngMidLen is deployed as the baseline model. 
It performs comparatively on KAIST-Sem-LS while provides much worse result on KITTI-Sem-LS.
Sem-LSD\_v0 model substitutes AngMidLen by LineAsObj and adopts regression loss to learn direction. It performs much more robust on both datasets.
By using $L_{d_{cls}}$ to learn direction (column ``D-Cls'' in the table), we build Sem-LSD\_v1 model which consistently outperforms Sem-LSD\_v0 for around $3\%$ on $\mathrm{mAP}@0.5$.
Two implementations of BAM (BAM51 and BAM33) works differently on two datasets. 
On KAIST-Sem-LS, BAM33 achieves the best performance while BAM51 performs similarly as Sem-LSD\_v1. On the other hand, BAM51 slightly outperforms Sem-LSD\_v1 on KITTI-Sem-LS while BAM33 performs worse. 
%Such results are expected. As KAIST-Sem-LS captures the view of pure city streets while KITTI-Sem-LS is mainly from Europe urban areas, the appearance of Sem-LS from KAIST-Sem-LS is less diverse.
%BAM can helps to better handle vertical and horizontal Sem-LSs.
Our proposed Sem-LSD is demonstrated to provide effective detections on two benchmarks.
However, the extreme variations in Sem-LS scales and angles (such as remote building and nearby curb) still remain challenging.

%\begin{table*}[!t]
%	\begin{center}
%		\begin{tabular}{c|c|c|c|c}
%			\hline
%			\multirow{2}{*}{Models} & \multicolumn{2}{c}{KAIST-Sem-LS} & \multicolumn{2}{c}{KITTI-Sem-LS} \\
%			\cline{2-5}
%			 & $\mathrm{mAP}@0.5$ & $\mathrm{mAP}_3@0.5$ & $\mathrm{mAP}@0.5$ & $\mathrm{mAP}_3@0.5$ \\
%			\hline\hline
%			%LSD    & 0.0086 & --- \\
%			%\hline
%			AngMidLen & 0.4742 & 0.6605 & 0.2607 & 0.6434 \\
%			\hline
%			LineAsObj + Reg & 0.4559 & 0.6684 & 0.4103 & 0.6529 \\
%			\hline
%			LineAsObj + Cls & 0.4837 & 0.6591 & 0.4533 & \textcolor{red}{0.6680} \\
%			\hline
%			LineAsObj + Cls + BAM33 & \textcolor{red}{0.5002} & \textcolor{red}{0.6805} & 0.4304 & 0.6494 \\
%			\hline
%			LineAsObj + Cls + BAM51 & 0.4841 & 0.6562 & \textcolor{red}{0.4670} & 0.6532 \\
%			\hline
%		\end{tabular}
%	\end{center}
%	\caption{Results from Sem-LSD using different BAM with dla34 backbone on KITTI-Sem-LS dataset.}
%	\label{tab:Sem-LSD-ablation}
%\end{table*}

%\begin{figure}[!th]
%	\begin{center}
%		\hspace{0cm}
%		\includegraphics[width=0.8\linewidth]{figures/DetRes.jpg}
%	\end{center}
%	\vspace{-10pt}
%	\caption{Sample results from LSD and different Sem-LSD with BAMs. (a) ground truth labels; (b) LSD results; (c) Sem-LSD\_v1; (d) BAM51; (e) BAM33}
%	\label{fig:det_res}
%\end{figure}

%\noindent\textbf{Sem-LSD efficiency.} $\;$ 
Efficiency is another important factor since many robotics and autonomous driving applications requires real-time processing. 
We evaluate the efficiency of Sem-LSD by comparing runtime on a single GTX 1080Ti with i7 CPU and a Nvidia Jetson TX2 platform \cite{platform_tx2}.
All tests are implemented with PyTorch \cite{framework_pytorch}. Results are shown in Table \ref{tab:Sem-LSD-KITTI-speed}.
Four backbones are tested on KAIST-Sem-LS, namely, ResNet-18 \cite{network_resnet}, DLA-34, Hourglass-104 \cite{network_stackedhourglass}, and ResNeXt-101 \cite{network_resnext}. ResNet-18 has the lowest computational complexity but also the lowest $\mathrm{mAP}@0.5$. It runs at above $164$ fps (6.1ms/image). Hourglass-104 backbone provides the best performance at the cost to run only with $36$ fps (27.6ms/image). DLA-34\_TX2 is a compatible version on Jetson TX2 platform with deformable convolution replaced by an ordinary convolution. It has an $5\%$ drop on $\mathrm{mAP}@0.5$ compared to the original DLA-34. We port DLA-34\_TX2 to TX2 platform with ONNX \cite{onnx} format achieving a $7.5$ fps with TensorRT acceleration, which is good enough for near-real time processing on end device. 

\begin{table}[!ht]
	\begin{center}
		\begin{tabular}{c|c|c}
			\hline
			%\multirow{2}{*}{Models} & $\mathrm{mAP}$ & $\mathrm{mAP}_3$ & \multirow{2}{*}{fps} \\
			% & $@0.5$ & $@0.5$ & \\
			Models & $\mathrm{mAP}@0.5$ & fps \\
			\hline\hline
			ResNet-18        & 0.4713 &  164 \\
			\hline
			DLA-34 				& 0.4837 &  51  \\
			\hline
			ResNeXt-101 & 0.5031 & 21 \\
			\hline
			Hourglass-104 & 0.5180 & 36 \\
			\hline\hline
			DLA-34\_TX2     & 0.4320 &  7.5 \\
			\hline
		\end{tabular}
	\end{center}
	\vspace{-5pt}
	\caption{Sem-LSD with different backbone tested on KAIST-Sem-LS dataset.}
	\label{tab:Sem-LSD-KITTI-speed}
\end{table}

%\vspace{2pt}\noindent\textbf{Compare Sem-LSD and other line segment detectors.}$\;$
%We show the advantages of Sem-LSD over existing line-shaped element detectors qualitatively and quantitatively. 

%In Figure \ref{fig:diff_lines_exp}, we show detection results of LSD (a and c) and Sem-LSD (b and d) from two sets of continuous keyframes (i1)-(i3) and (ii1)-(ii3). Five cases in (i) are highlighted to demonstrate the advantage of Sem-LSD. Case A, C, and D are complete edges. The pole in A has background with similar color and the pole in C is blocked by a traffic sign. The left-side of building roof in D is occluded behind some trees. 
%LSD failed to retrieve complete line segments in these cases. 
%Case B is a building edge where LSD provides in-consistency results across three continuous keyframes, which can be miss-leading in applications such as image matching. 
%Last but not the least, case E is a pole with weak pixel-level gradients but strong semantic meaning. LSD misses it while Sem-LSD provides consistency detections.
%Similar cases can be noticed in (ii). 
%(ii) also illustrate that Sem-LSD can easily avoid unstable regions such as vehicles by semantics. 

%\begin{figure}[!t]
%	\begin{center}
%		\hspace{0cm}
%		\includegraphics[width=1.0\linewidth]{figures/cont_frames_v1.png}
%	\end{center}
%	\vspace{-10pt}
%	\caption{Detection results of LSD (a and c) compared with Sem-LSD (b and d) from two sets (i and ii) of three continuous keyframes (1-3). In (i) keyframes, five cases (A-E) are highlighted to show preferred features of Sem-LSD.}
%	\label{fig:diff_lines_exp}
%\end{figure}

\begin{figure}[!t]
	\begin{center}
		\hspace{0cm}
		\includegraphics[width=1.0\linewidth]{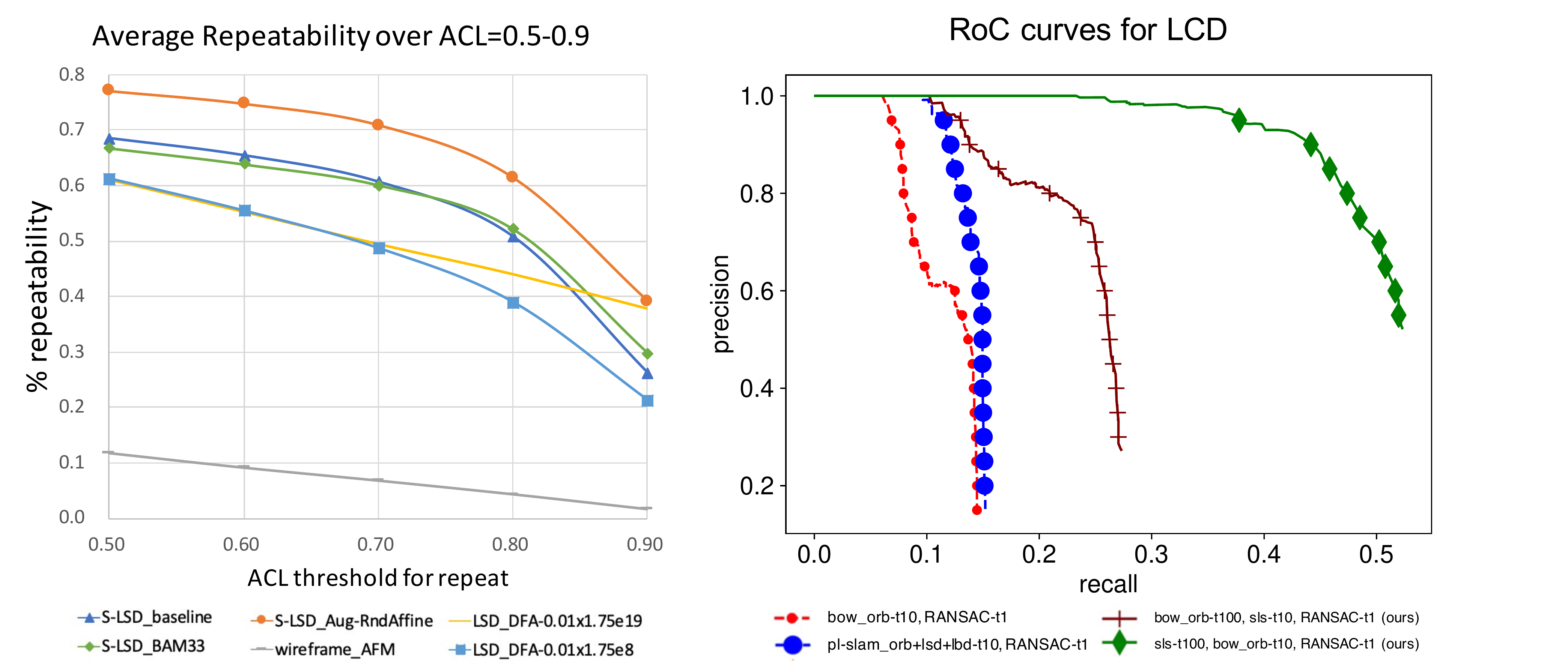}
	\end{center}
	\vspace{-10pt}
	\caption{Left: Curves of averaged repeatability over different ${th}_{re}$ for different detectors and their variations. Right: RoC curve for LCD on KAIST URBAN dataset.}
	\label{fig:repeatability}
\end{figure}

\vspace{2pt}\noindent\textbf{Experiment on repeatability.}$\;$
To further demonstrate the robustness of Sem-LS quantitatively, we design an experiment to compare the repeatability of Sem-LSD, LSD, and one of the state-of-the-art wireframe parsers, AFM \cite{lineshaped_afm}. 
Repeatability measures the ratio of repeated detections of the same local feature visible from multiple images. Line segment correspondence is required to calculate repeatability.
However, as Sem-LS is designed to retrieve robust semantics instead of low-level features, we intentionally avoid matching Sem-LS by descriptors such as LBD \cite{lineshaped_lbd} which can break the semantical robustness especially under conditions such as different illuminations and seasons. 
It is also un-realistic to manually label the correspondence of LSD results and wireframes.
Hence, we adopt random affine transformation to generate correspondence between original images and their transformed counterparts. 
By such known correspondence, we evaluate repeatability of the three detectors on our KAIST-Sem-LS test set. 
An image $I$ is randomly transformed to generate a matched image pair $(I, I^t)$. Repeatability $\mathrm{Re}$ is calculated by:
\begin{equation}
\mathrm{Re} = \frac{N_{I\rightarrow I^t} + N_{I^t\rightarrow I}}{N_I + N_{I^t}},
\end{equation}
where $N_{I\rightarrow I^t}$ and $N_{I^t\rightarrow I}$ are the amount of repeated detections from $I$ to $I^t$ and from $I^t$ to $I$, respectively. $N_I$ and $N_{I^t}$ are total number of detections from image pair $(I, I^t)$.
A line segment is said to ``repeat'' if it has an ACL value larger than ${th}_{re}$ with its ground truth match after transformation. We calculate $Re$ with five ${th}_{re} = [0.5, 0.6, 0.7, 0.8, 0.9]$ and report the mean averaged repeatability $mARe$. 
Results are shown in Table \ref{tab:repeatability}.
We implement LSD with two different false positive rejection thresholds at $0.01\times1.75e8$ and $0.01\times1.75e19$. Sem-LSD\_Aug-RndAffine is a customized Sem-LSD trained with data augmentation using random affine transformation. Curves of $Re$ with different ${th}_{re}$ are shown in the left plot of Figure \ref{fig:repeatability}.
It is clear that Sem-LSD achieves the best performance. Sem-LSD\_v1 leads LSD by around $5\%$. Sem-LSD\_Aug-RndAffine further improves the repeatability by another $10\%$. 
%It is worth noting that such performance is achieved given only around $50\%$ mAP of Sem-LSD. 

\begin{table}[!ht]
	\begin{center}
		\begin{tabular}{c|c}
			\hline
			%\multirow{2}{*}{Models} & $\mathrm{mAP}$ & $\mathrm{mAP}_3$ & \multirow{2}{*}{fps} \\
			% & $@0.5$ & $@0.5$ & \\
			Detector & $\mathrm{mARe}$ \\
			\hline\hline
			AFM & 6.79\% \\
			\hline
			LSD\_DFA-$0.01\times1.75e8$ & 45.08\% \\
			\hline
			LSD\_DFA-$0.01\times1.75e19$ & 49.56\% \\
			\hline
			Sem-LSD\_v1 & 54.34\% \\
			\hline
			Sem-LSD\_BAM33 & 54.53\% \\
			\hline
			Sem-LSD\_Aug-RndAffine & 64.64\% \\
			\hline
		\end{tabular}
	\end{center}
	\vspace{-5pt}
	\caption{Sem-LSD with different backbone tested on KAIST-Sem-LS dataset.}
	\label{tab:repeatability}
\end{table}

\vspace{2pt}\noindent\textbf{An example application of Sem-LS in large-scale LCD.}$\;$
In order to demonstrate the potential of Sem-LS in real-world applications and to show the sparsity of Sem-LS is sufficiently dense as image representation, we develop a Sem-LS-based LCD algorithm and test it on KAIST URBAN dataset with sampled query images from the 28th sequence and database images from the 39th sequence.
Based on the definition of Sem-LS, traditional descriptor-based matching such as LBD \cite{lineshaped_lbd} is not suitable and can even break the robustness under changing appearance. Hence, we apply a straight-forward voting mechanism to match Sem-LS based on their semantics and geometrics. 
We compared recalls under three precision thresholds, namely $99\%$ (R@P0.99), $95\%$ (R@P0.95), and $80\%$ (R@P0.80). RANSAC \cite{LC_ransac} is applied in all approaches to determine the top match. Results are shown in Table \ref{tab:recalls}. Figure \ref{fig:repeatability} right plot is the RoC curve for LCD results.
Our Sem-LS based approach is shown to outperform classical BoW \cite{bovw_bow} with ORB \cite{descriptor_orb} and line segments based \cite{slam_plslam} approaches by wide margins. 
Due to the space limitation, detailed implementations and more experimentations results are presented in supplementary materials.

\begin{table}[!h]
	\begin{center}
		\hspace{-5mm}
		\begin{tabular}{c|c|c|c}
			\hline
			Model & R@P0.99 & R@P0.95 & R@P0.80 \\
			\hline
			bow\_orb-top10 & \multirow{1}{*}{0.0604} &  \multirow{1}{*}{0.0686} &  \multirow{1}{*}{0.0792} \\
			%RANSAC-t1 & & & \\
			\hline
			pl-slam\_orb-top10 & \multirow{1}{*}{0.1013} &  \multirow{1}{*}{0.1127} &  \multirow{1}{*}{0.1315} \\
			%RANSAC-t1 & & & \\
			\hline
			bow\_orb-top100 & \multirow{2}{*}{0.1021} & \multirow{2}{*}{0.1283} & \multirow{2}{*}{0.2092} \\
			Sem-LS-top10 & & & \\
			\hline
			Sem-LS-top100 & \multirow{2}{*}{\textbf{0.2614}} & \multirow{2}{*}{\textbf{0.3775}} & \multirow{2}{*}{\textbf{0.4681}} \\
			bow\_orb-top10 & & & \\
			%RANSAC-t1 (ours) & & & \\
			\hline
		\end{tabular}
	\end{center}
	\vspace{-5pt}
	\caption{Comparisons of LCD recalls under $0.99$, $0.95$, and $0.8$ precisions. }
	\label{tab:recalls}
\end{table}

\begin{figure}[!th]
	\begin{center}
		\hspace{0cm}
		\includegraphics[width=0.9\linewidth]{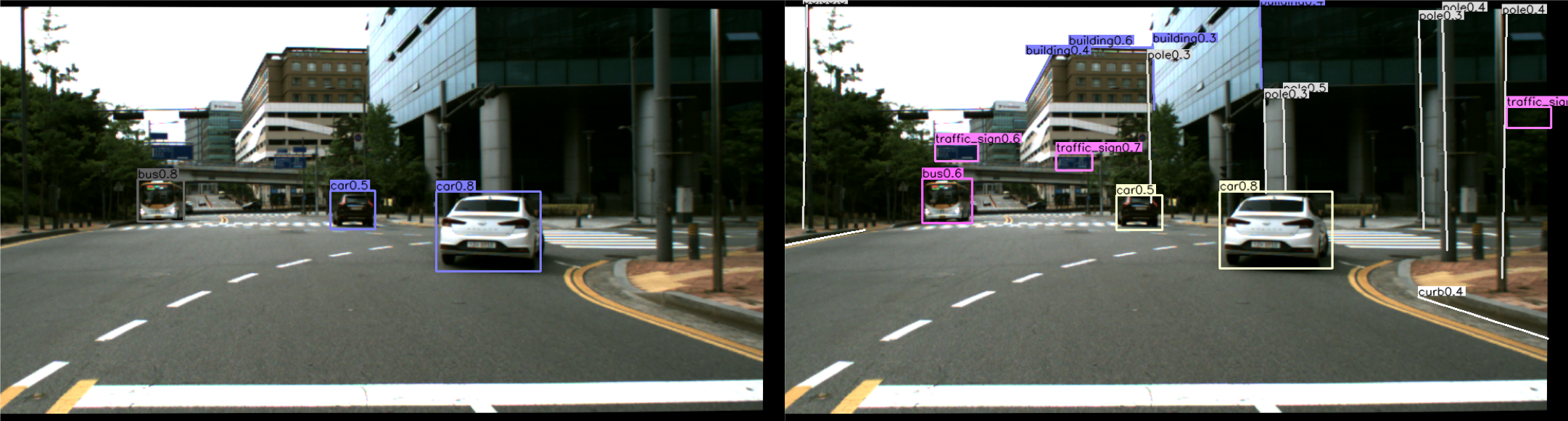}
	\end{center}
	\vspace{-10pt}
	\caption{Top: object detector result. Bottom: unified detector result, where some missed small objects from object detector are successfully detected.}
	\label{fig:objVSobjline}
\end{figure}

\vspace{2pt}\noindent\textbf{Unified detector.}$\;$
We evaluate Sem-LSD as a unified detector for both objects and Sem-LS based on LineAsObj as a general encoding.
%As we have described in Section \ref{sec:representation}, LineAsObj enables a general encoding of both Sem-LS and objects. Hence, we approach one step further to try implement Sem-LSD as a unified detector for both objects and Sem-LSs. 
Seven types of objects are labeled including person, car, bus, traffic sign, traffic light, road light, and road light area on KAIST-Sem-LS for a total of $2490$ images, where $2072$ for training and the rest $418$ for testing. 
We use Hourglass-104 as the backbone for all three detectors. $L_{d_{cls}}$ is applied to learn directions. 
We show our results in Table \ref{tab:Sem-LSD-objline}. 
Train labels and test labels denote the data annotations used to train and test the detector. 
%We show $\mathrm{mAP}@0.5$ results. 
We compare unified detector (Uni-Sem-LSD) with single-tasked detectors, namely, Sem-LSD and ObjDetector, to see if Uni-Sem-LSD is able to retain similar level of performance. 
In terms of Sem-LS detection, Sem-LSD has $\mathrm{mAP}@0.5=0.4948$ whereas Uni-Sem-LSD achieves $0.4860$.
On the other hand, ObjDetector surpasses Uni-Sem-LSD with $0.5519$ over $0.5402$.
We also notice that Uni-Sem-LSD better retrieves small objects than ObjDetector. 
Figure \ref{fig:objVSobjline} shows an example that some small objects are missed by ObjDetector but successfully detected by Uni-Sem-LSD.
Such initial results demonstrates the capability of Uni-Sem-LSD using LineAsObj. It achieves similar results on both tasks while doubling the inference efficiency.

\begin{table} [!h]
	\begin{center}
		\begin{tabular}{c|c|c|c}
			\hline
			Model & Train labels & Test labels & $\mathrm{mAP}@0.5$\\
			\hline\hline
			Uni-Sem-LSD & obj-line & obj-line & 0.5093  \\
			\hline
			Uni-Sem-LSD & obj-line	& obj & 0.5402 \\
			\hline
			ObjDetector & obj	& obj & 0.5519 \\
			\hline
			Uni-Sem-LSD & obj-line & line & 0.4860 \\
			\hline
			Sem-LSD & line & line & 0.4948 \\
			\hline
		\end{tabular}
	\end{center}
	\vspace{-5pt}
	\caption{Test Sem-LSD as a unified detector for both object and Sem-LS.}
	\label{tab:Sem-LSD-objline}
\end{table}

%------------------------------------------------------------------------
\section{Conclusion} \label{sec:conclusion}
In this work, we propose a new line-shaped image representation Sem-LS combines high-level stable semantics to visually salient line segments. 
We argue that Sem-LS is more robust against existing line-shaped representations under challenging conditions such as changing illuminations, seasons, and occlusions. 

To solve the Sem-LS detection problem, we propose two new datasets, a real-time detector Sem-LSD, customized BAM, and ACL for evaluation. 
We evaluate both the efficacy and efficiency of Sem-LSD and also develop a unified detector illustrating the possibility of solving Sem-LS and object detection simultaneously. It is demonstrated that Sem-LS detection problem is a challenging and valid research topic.
Sem-LSD is also shown to provide more robust and consist detections compared with LSD and wireframe by outperforming $15\%$ in terms of repeatability.
An example implementation on large-scale LCD problem further demonstrates the potential of Sem-LS in real-world applications.

%The datasets and codes are made publicly available to better benefit the community. 
%We also show that Sem-LSD provides more robust and consist detections compared with LSD and wireframe by outperforming 15\% in terms of repeatability. 

%We demonstrate that Sem-LS is an important visual element. Its detection problem is a challenging and valid research topic. The datasets and codes are made publicly available to better benefit the community. 

{\small
\bibliographystyle{ieee_fullname}
\bibliography{SLSD}
}

\end{document}